\documentclass[conference]{IEEEtran}
\ifCLASSINFOpdf
	\usepackage{graphics} 
	\usepackage{epsfig} 
	\usepackage{mathptmx} 
	\usepackage{times} 
	\usepackage{amsmath} 
	\usepackage{amssymb}  
	\usepackage{textcomp}
	\usepackage{multirow}
	\usepackage{colortbl}
	\usepackage{xcolor}
	\usepackage{hhline}
	\usepackage{subfigure}
	\usepackage{bm}
	\usepackage{algorithm}
	\usepackage{algpseudocode}
	\graphicspath{{}}
\else
\fi
\begin{document}
%

\title{GridSim: A Vehicle Kinematics Engine for Deep Neuroevolutionary Control in Autonomous Driving}


%
\author{Bogdan~Tr\u{a}snea,
Liviu~A.~Marina,
Andrei Vasilcoi,
Claudiu~R.~Pozna and
Sorin~M.~Grigorescu \\
Elektrobit Automotive Romania, Transilvania University of Brasov, 500036 Romania \\
Email: bogdan.trasnea@unitbv.ro}

\markboth{IEEE TRANSACTIONS ON ROBOTICS}%
{Shell \MakeLowercase{\textit{et al.}}: Bare Demo of IEEEtran.cls for IEEE Journals}
%

\maketitle

\begin{abstract}
Current state of the art solutions in the control of an autonomous vehicle mainly use supervised end-to-end learning, or decoupled perception, planning and action pipelines. Another possible solution is deep reinforcement learning, but such a method requires that the agent interacts with its surroundings in a simulated environment. In this paper we introduce GridSim, which is an autonomous driving simulator engine running a car-like robot architecture to generate occupancy grids from simulated sensors. We use GridSim to study the performance of two deep learning approaches, deep reinforcement learning and driving behavioral learning through genetic algorithms. The deep network encodes the desired behavior in a two elements fitness function describing a maximum travel distance and a maximum forward speed, bounded to a specific interval. The algorithms are evaluated on simulated highways, curved roads and inner-city scenarios, all including different driving limitations.

\end{abstract}


%
\IEEEpeerreviewmaketitle

\section{Introduction and related work}
\label{sec:introduction}
\begin{figure}
	\centering
	\begin{center}
		\includegraphics[scale=0.3]{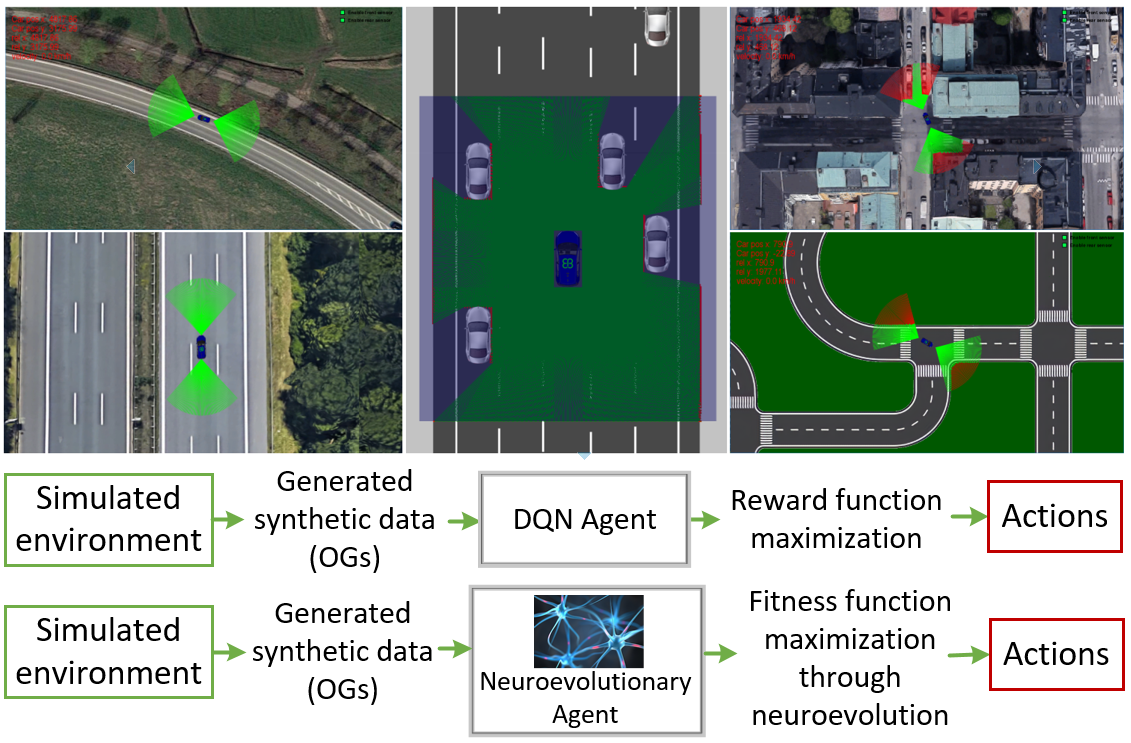}
		\caption{\textbf{GridSim and two possible pipelines for deep neural control of a simulated car.} (top) GridSim driving scenarios. (middle) DQN Agent pipeline using the input OGs for interacting with the simulated environment in order to maximize its reward function. (bottom) Neuroevolutionary Agent: the DNN's weights are evolved using genetic algorithms with altered breeding rules, in order to maximize a two elements fitness function.}
        \label{fig:gridsim}
	\end{center}
\end{figure}

%
%
%
%
\IEEEPARstart{T}{he} ability of an autonomous car to navigate without human input has become a mainstream research topic in the quest for autonomous driving. In this paper we propose a simulated environment engine for learning autonomous driving behaviors, entitled GridSim (see Fig.~\ref{fig:gridsim}). The simulator uses an Occupancy Grid (OG) sensor model for interacting with the simulated environment. As shown in Fig.~\ref{fig:gridsim}, we used GridSim synthetic information to train two types of learning algorithms commonly used in Autonomous Driving: a Deep Q-Network (DQN Agent)~\cite{sallab2017deep} and a Deep Neuroevolutionary Agent evolved via an evolutionary training method~\cite{UberDeepNeuroevolution2018}.

An AV must be able to sense its own surroundings and form an environment model consisting of moving and stationary objects~\cite{survey_comp_vision_AV}, and to further use this information in order to learn long term driving strategies. These driving policies govern the vehicle's motion~\cite{Survey_self_driving} and automatically output control signals for steering wheel, throttle and brake. Reinforcement learning has been applied to a wide variety of robotics related tasks, such as robot locomotion and autonomous driving~\cite{DRL_ScalableDL_2018}. However, DRL requires the agent to interact with its environment. The reward is used as a pseudo label for training a DNN which is then used to estimate an action-value function, also known as a Q-value function, for approximating the next best driving actions. This is in contrast to end2end learning, where labeled training data has to be provided. The DeepTraffic competition from MIT~\cite{fridman2018deeptraffic} is a good example of RL in simulated environments, because the control system for the ego-car is handled by a DQN agent, which uses a discrete occupancy grid as a simplified representation of the environment.

\section{Simulation engine, methods and experiments}
\label{sec:sim_methods_experiments}

The developed simulation engine is coined GridSim and it is described as an autonomous driving simulator which uses the non-holonomic robot car kinematics~\cite{kong2015kinematic}. The steering is modelled through angle $\delta$ as an extra degree of freedom on the front wheel, while the "non-holonomic" assumption is expressed as a differential constraint on the motion of the car, which restricts the vehicle from making lateral displacements, without simultaneously moving forward.

The simulator has been developed from scratch, using Python and PyGame library, in order to support development and validation of autonomous driving systems, and it can be used in both CPU and GPU configurations. Snapshots of GridSim's possible scenarios can be seen at the top of the Fig.~\ref{fig:gridsim}. The simulated sensors have a field of view (FOV) of 120 degrees, and they react when an obstacle is sensed, by marking it as an occupied area. The dynamic obstacles are represented by traffic cars, which have their trajectory randomly generated from a uniform distribution of the possible free spaces inside the given scenario. We use GridSim to study the performance of two simulation based autonomous driving approaches: deep reinforcement learning and the control of a deep neuroevolutionary agent.

\begin{figure}
	\centering
	\begin{center}
		\includegraphics[scale=0.59]{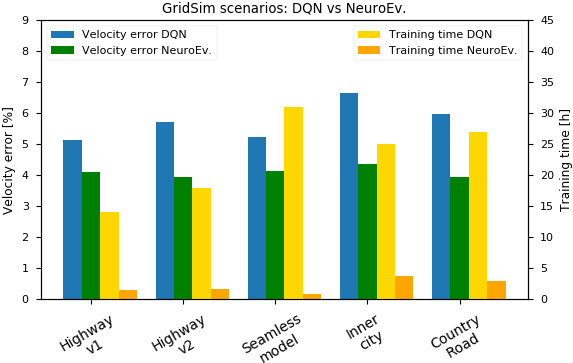}
		\caption{\textbf{Comparison of the DQN and Neuroevolutionary agents in the five GridSim scenarios.} Performance comparison in regards to overall velocity error percentage and average training time of both models with different scenarios in GridSim. We observe that the velocity error of the neuroevolutionary approach is smaller in all scenarios, while keeping its training time low.}
        \label{fig:comparison_ga_dqn}
	\end{center}
\end{figure}

The neuroevolutionary part of the algorithm represents the evolution of the weights of a deep neural network by using a population-based genetic algorithm, with altered breeding rules (custom tournament selection). The training is performed against a multi-objective fitness function which maximizes two elements: the traveled path and the longitudinal speed. This learning procedure was first proposed by the authors for training a generative one-shot learning classifier~\cite{SorinGr_ICRA2018}. It aims to compute optimal weights for a collection of $K$ deep networks:
\begin{equation}
	\bm{\varphi}(\cdot; \bm{\Theta}) = \begin{bmatrix} \varphi_1(\cdot; \Theta_1), ..., \varphi_i(\cdot; \Theta_i), ..., \varphi_K(\cdot; \Theta_K), \end{bmatrix}^T
\end{equation}

The weights of a single deep network are stored in a so-called solution vector $\Theta = \begin{bmatrix} \theta_1, \theta_2, ..., \theta_n, \end{bmatrix}^T$, composed of $n$ decision variables $\theta_i$, with $i = 1, ..., n$ and $\theta \in \mathbb{R}^n$. $\theta_i$ represents a weight parameter in a single network. The agent controls the ego-car using the elite individual DNN from the given generation, while the custom tournament selection algorithm ensures that the best accuracy individuals carry on to the next generation unmodified.

As a comparison to our Neuroevolutionary Agent, we have implemented a DQN agent, which uses a decision space of eight actions. The algorithm starts from an initial state and proceeds until the the agent has collided with its surroundings. In every step, the agent is described by the current state $s$, it follows policy $p(s)$ and observes the next state together with the reward received from the environment. The reward policy is constructed in the following way:

\begin{equation}
	\rho_t=f(\delta_t)*0.15+f(v_t)*0.15+f(S_t)*0.7  
	\label{eq:reward_policy_dqn}
	\end{equation} 
	\begin{equation}
	S = \begin{bmatrix} s_1, s_2, ..., s_n \end{bmatrix}^T
	\label{eq:sensor_vector}
	\end{equation} 
	\begin{equation}
	f(S) = min(S)
	\label{eq:sensor_policy}
\end{equation}

\begin{figure}
	\centering
	\begin{center}
		\includegraphics[scale=0.31]{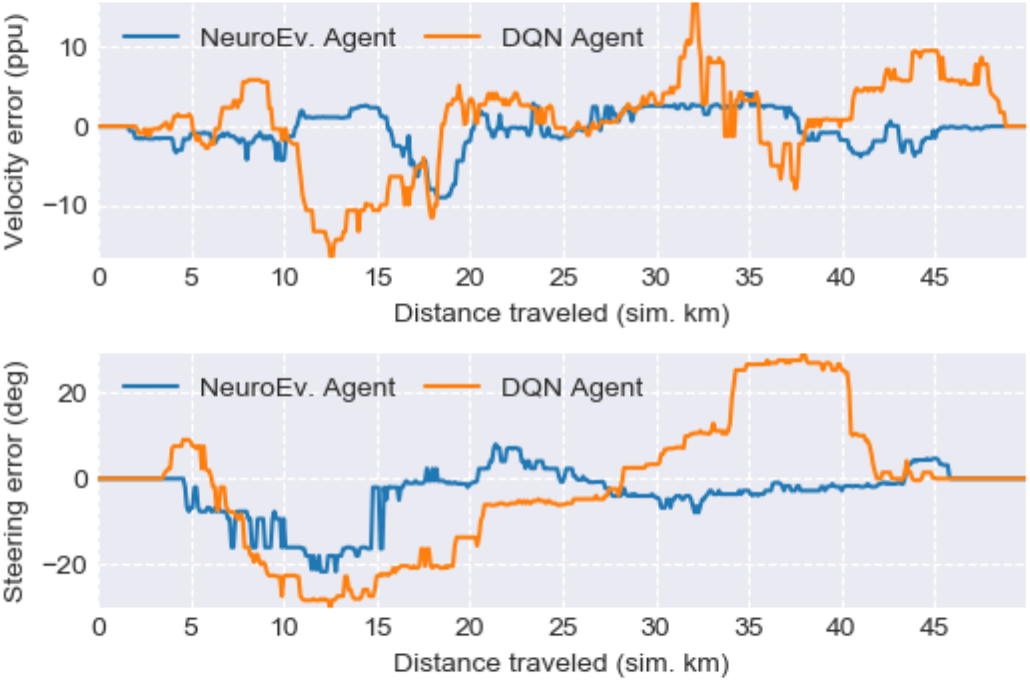}
		\caption{\textbf{Seamless generated model scenario velocity and steering errors when compared with human control reference}. The absolute values obtained with the Neuroevolutionary Agent: $mean(e_{v_f}) = 1.692 ppu$, $max(e_{v_f}) = 8.988 ppu$, ($mean(e_{\delta}) = 4.726 deg$, $max(e_{\delta}) = 21.619 deg$. The neuroevolutionary approach delivers a higher accuracy when compared to the DQN network, which is tailored to operate on simulation data.}
        \label{fig:vel_steering_errors}
	\end{center}
\end{figure}

\noindent where $\rho$ is the total normalized reward, $f(\delta)$ is the distance travelled, $f(v)$ is the current velocity of the vehicle, $f(S)$ is the sensor policy and $S$ is the sensor action-value vector. The algorithm continues until the convergence of the Q function, or until a certain number of episodes is reached, while also ensuring a sanity check of $15$ actions.

\section{Conclusion and future work}
\label{sec:conclusion}

In summary, this paper presents two possible approaches for controlling a simulated kinematic model, namely the Neuroevolutionary and DQN Agents. For this purpose, we have implemented GridSim, which is a multiple-scenario, two dimensional birds' eye view driving simulator. We have reached the conclusion that the genetic evolution of a deep network's weights converges faster and achieves better results than the DQN Agent in the virtual driving scenarios in which the algorithms were tested. As future work, we would like to extend the control algorithms for the simulated car, together with using a more complex kinematic model.


%

\ifCLASSOPTIONcaptionsoff
  \newpage
\fi



%

\bibliographystyle{IEEEtran}
\bibliography{references}

%








\end{document}